\documentclass[10pt,twocolumn,letterpaper]{article}

\usepackage{wacv}
\usepackage{times}
\usepackage{epsfig}
\usepackage{graphicx}
\usepackage{amsmath}
\usepackage{amssymb}
\usepackage{nccmath}
\usepackage{xcolor}
\usepackage{color, colortbl}

\def\wacvPaperID{****} %

\wacvfinalcopy %
\ifwacvfinal
\def\assignedStartPage{9876} %
\fi

\newcommand\method{CenterFusion}
\definecolor{ForestGreen}{RGB}{34,139,34}
\definecolor{LightCyan}{rgb}{0.88,1,1}
\definecolor{Gray}{gray}{0.8}

\ifwacvfinal
\usepackage[breaklinks=true,bookmarks=false]{hyperref}
\else
\usepackage[pagebackref=true,breaklinks=true,colorlinks,bookmarks=false]{hyperref}
\fi

\ifwacvfinal
\else
\fi

\begin{document}

\title{CenterFusion: Center-based Radar and Camera Fusion for 3D Object Detection}

\author{Ramin Nabati, Hairong Qi\\
University of Tennessee Knoxville\\
{\tt\small \{rnabati, hqi\}@utk.edu}
}

\maketitle

\begin{abstract}
   The perception system in autonomous vehicles is responsible for detecting and 
   tracking the surrounding objects. This is usually done by taking advantage of 
   several sensing modalities to increase robustness and accuracy, which makes 
   sensor fusion a crucial part of the perception system. In this paper, we focus on 
   the problem of radar and camera sensor fusion and propose a middle-fusion approach 
   to exploit both radar and camera data for 3D object detection. Our approach,
   called \method{}, first uses a center point detection network to detect objects 
   by identifying their center points on the image. It then solves the key data association problem using a novel frustum-based 
   method to associate the radar detections to their corresponding object's center 
   point. The associated 
   radar detections are used to generate radar-based feature maps to complement the 
   image features, and regress to object properties such as depth, rotation and 
   velocity. We evaluate CenterFusion on the challenging nuScenes dataset, where it improves 
   the overall nuScenes Detection Score (NDS) of the state-of-the-art camera-based 
   algorithm by more than 12\%. We further show that CenterFusion significantly
   improves the velocity estimation accuracy without using any additional 
   temporal information. The code is available at 
   \href{https://github.com/mrnabati/CenterFusion}{https://github.com/mrnabati/CenterFusion}.
   
   \end{abstract}
   
   \section{Introduction}
   Autonomous vehicles are usually equipped with different types of sensors 
   to take advantage of their complimentary characteristics.
   Using multiple sensor modalities increases robustness and accuracy, but also
   introduces new challenges in designing the perception system. Sensor fusion is 
   one of these challenges, which has motivated many studies in 2D and 3D object
   detection \cite{Chen_2017, Ku_2018,Liang_2019_CVPR,Nabati_2019}, semantic
   segmentation \cite{7139439,Meyer_2019} and object tracking
   \cite{7795718,fang2019camera} in recent years.
   
   Most of the recent sensor fusion methods focus on exploiting LiDAR and camera
   for 3D object detection. LiDARs use the time of flight of laser light pulses 
   to calculate distance to surrounding objects. LiDARs provide accurate 3D 
   measurement at close range, but the resulting point cloud becomes sparse at 
   long range, reducing the system's ability to accurately detect far away objects.
   Cameras provide rich appearance features, but are not a good source of 
   information for depth estimation. These complementary features have made 
   LiDAR-camera sensor fusion a topic of interest in recent years. 
   This combination has been proven to achieve high accuracy in 3D 
   object detection for many applications including autonomous driving,
   but it has its limitations. Cameras and LiDARs are both 
   sensitive to adverse weather conditions (e.g. snow, fog, rain), which can
   significantly reduce their field of view and sensing capabilities. 
   Additionally, LiDARs and cameras are not capable of detecting objects' 
   velocity without using temporal information. Estimating objects' 
   velocity is a crucial requirement for collision avoidance in many scenarios, 
   and relying on the temporal information might not be a feasible solution 
   in time-critical situations.
   
   \begin{figure*}[t]
      \includegraphics[width=0.99\textwidth]{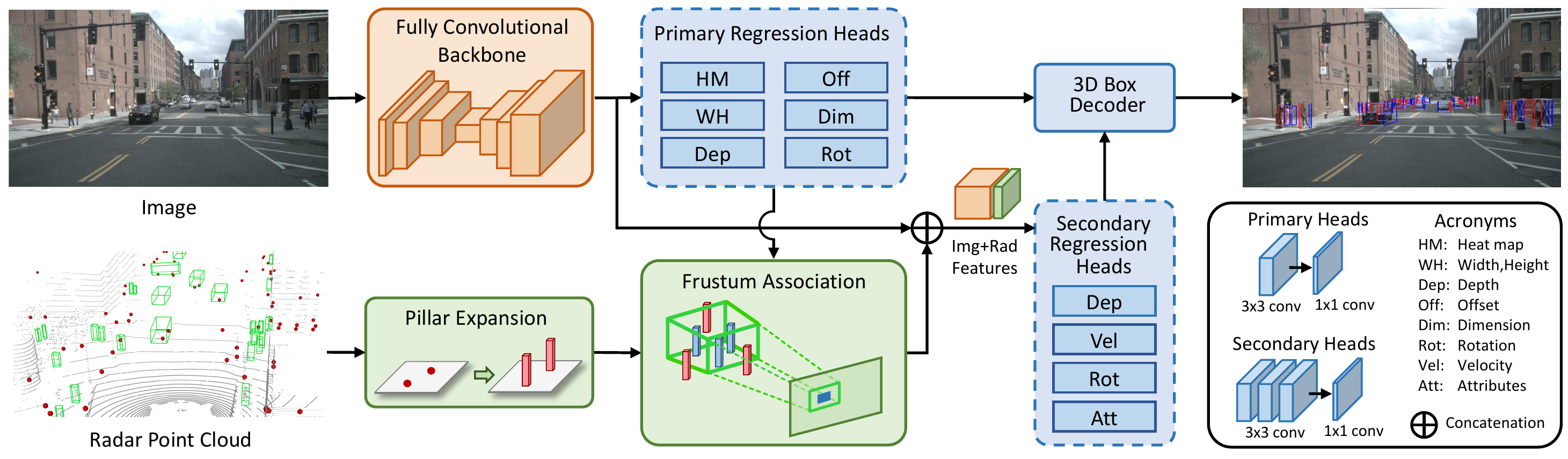}
      \caption{\method{} network architecture. Preliminary 3D boxes 
      are first obtained using the image features extracted by the backbone. The 
      frustum association module uses the preliminary boxes to associate 
      radar detections to objects and generate radar feature maps. 
      The image and radar features maps are then concatenated and used to 
      refine the preliminary detections by recalculating depth and rotation
      as well as estimating objects' velocity and attributes.}
      \label{fig:netArch}
   \end{figure*}

   For many years, radars have been used in vehicles for Advanced Driving
   Assistance System (ADAS) applications such as collision avoidance and 
   Adaptive Cruise Control (ACC). Compared to LiDARs and cameras, radars are very 
   robust to adverse weather conditions and are capable of detecting objects 
   at very long range (up to 200 meters for automotive radars). 
   Radars use the Doppler effect to accurately estimate the 
   velocities of all detected objects, without requiring any temporal information.
   Additionally, compared to LiDARs, Radar point clouds require less processing 
   before they can be used as object detection results.
   These features and their lower cost compared to LiDARs, 
   have made radars a popular sensor in autonomous driving applications. 
   
   Despite radar's popularity in the automotive industry, few studies 
   have focused on fusing radar data with other sensors. 
   One reason for this is the fact that there are not many datasets containing radar
   data for autonomous driving applications, which makes conducting research in 
   this area difficult. Additionally, due to inherent differences between LiDAR 
   and radar point clouds, applying or adapting existing LiDAR-based algorithms 
   to radar point cloud proves to be extremely difficult.
   Radar point clouds are significantly more sparse than their LiDAR counter parts,
   making it unfeasible to use for extracting objects' geometry information. 
   Aggregating multiple radar sweeps increases the density of the points, but also
   introduces delay into the system.
   Moreover, although radar point clouds are usually represented as points
   in the 3D coordinate system, the reported vertical measurement of the points 
   are usually not accurate or even non-existent, as most automotive radars only 
   report the distance and azimuth angle to objects.
   
   In order to effectively combine multiple sensing modalities, a variety of sensor 
   fusion schemes have been developed \cite{fengDeepMultimodalObject2020} taking 
   advantage of the hierarchical feature representation in neural networks.
   In an early fusion approach, the raw or pre-processed sensory data
   from different sensor modalities are fused together. With this approach,
   the network learns a joint representation from the sensing modalities. 
   Early fusion methods are usually sensitive to spatial or temporal misalignment
   of the data \cite{fengDeepMultimodalObject2020}.
   On the other hand, a late fusion approach combines the data from 
   different modalities at the decision level, and provides more 
   flexibility for introducing new sensing modalities to the network. However,
   a late fusion approach does not exploit the full potential 
   of the available sensing modalities, as it does not acquire the intermediate
   features obtained by learning a joint representation.
   A compromise between the early and late fusion approaches is referred to as 
   middle fusion. It extracts features from different modalities individually and 
   combines them at an intermediate stage, enabling the network to learn 
   joint representations and creating a balance between sensitivity and flexibility.

   We propose \method{}, a middle-fusion approach to exploit 
   radar and camera data for 3D object detection. \method{} focuses
   on associating radar detections to preliminary detection results obtained from 
   the image, then generates radar feature maps and uses it in addition to image features 
   to accurately estimate 3D bounding boxes for objects. Particularly, 
   we generate preliminary 3D detections using a key point detection network,
   and propose a novel frustum-based radar association method to accurately
   associate radar detections to their corresponding 
   objects in the 3D space. These radar detections are then mapped to the 
   image plane and used to create 
   feature maps to complement the image-based features. Finally, the fused
   features are used to accurately estimate objects' 3D properties
   such as depth, rotation and velocity. The network architecture for 
   CenterFusion is shown in Fig. \ref{fig:netArch}.
   
   We evaluate \method{} on the challenging nuScenes \cite{nuscenes2019} dataset, 
   where it outperforms all previous camera-based object detection methods in the
   3D object detection benchmark. We also show that exploiting radar information 
   significantly improves velocity estimation for objects without using any 
   temporal information.
   
   \section{Related Work}
   \subsection{Single-modality Methods}
   Monocular 3D object detection methods use a single camera to estimate 
   3D bounding boxes for objects. Many studies have been reported,  taking different approaches to extract the depth information
   from monocular images. 3D RCNN \cite{kundu3DRCNNInstanceLevel3D2018}
   uses Fast R-CNN \cite{girshickFastRCNN2015} with an additional head 
   and 3D projection. It also uses a collection of CAD models to learn class-specific
   shape priors for objects. Deep3DBox \cite{mousavian3DBoundingBox2017a} regresses a
   set of 3D object properties using a convolutional neural network first, then 
   uses the geometric constraints of 2D bounding boxes to produce a 3D 
   bounding box for the object. CenterNet \cite{zhou2019objects} takes a different approach and 
   uses a keypoint detection network to find objects' center point on the image. Other 
   object properties such as 3D dimension and location are obtained by regression
   using only the image features at the object's center point.
   
   LiDARs have been widely used for 3D object detection and tracking in autonomous
   driving applications in recent years. The majority of LiDAR-based methods either 
   use 3D voxels \cite{li3DFullyConvolutional2017,zhouVoxelNetEndtoEndLearning2017} 
   or 2D projections 
   \cite{liVehicleDetection3D2016, chenMultiView3DObject2017, yanSECONDSparselyEmbedded2018, yangPIXORRealtime3D2019} 
   for point cloud representation. Voxel-based methods are usually slow as a 
   result of the voxel grid's high dimensionality, and projection-based methods 
   might suffer from large variances in object shapes and sizes depending on the 
   projection plane. PointRCNN \cite{shiPointRCNN3DObject2019} directly 
   operates on raw point clouds and generates 3D object proposals in 
   a bottom-up manner using point cloud segmentation. These proposals are refined in 
   the second stage to generate the final detection boxes.
   
   \begin{figure}[t]
       \begin{center}
           \includegraphics[width=0.4\textwidth]{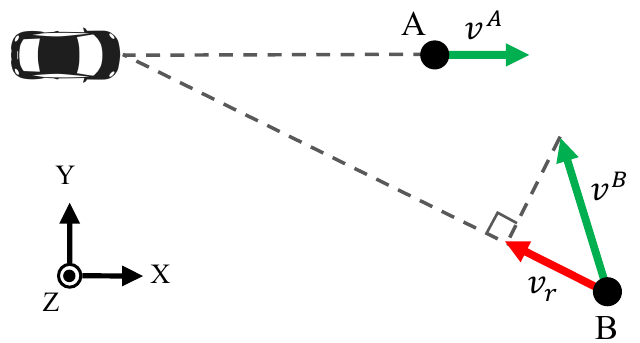}
           \caption{Difference between actual and radial velocity. For target A, 
           velocity in the vehicle coordinate system and the radial velocity 
           are the same ($v^A$). For target B on the other hand, radial velocity ($v_r$) 
           as reported by the radar is different from the actual velocity of the 
           object ($v^B$) in the vehicle coordinate system.}
           \label{fig:radarVel}
       \end{center}
   \end{figure}
   
   \subsection{Fusion-based Methods}
   Most existing sensor fusion methods focus on the LiDAR 
   and camera fusion problem. MV3D \cite{Chen_2017} extracts features from the
   front view and Bird's Eye View (BEV) representations of the LiDAR data, 
   in addition to the RGB image. The features obtained from the LiDAR's BEV are then used 
   to generate 3D object proposals, and a deep fusion network is used to combine 
   features from each view and predict the object class and box orientations.
   PointFusion \cite{xuPointFusionDeepSensor2018} processes the image and LiDAR 
   data using a CNN and a PointNet model respectively, and then generate 3D object 
   proposals using the extracted features. Frustum PointNet \cite{qiFrustumPointNets3D2018a}
   directly operates on the raw point clouds obtained from an RGB-D camera 
   and uses the RGB image and a 2D object detector to localize objects in the point cloud.
   
   Few studies have focused on fusing radars with other sensors
   for autonomous driving applications. RadarNet \cite{yangRadarNetExploitingRadar2020}
   fuses radar and 
   LiDAR data for 3D object detection. It uses an early fusion mechanism to learn 
   joint representations from the two sensors, and a late-fusion mechanism to 
   exploit radar's radial velocity evidence and improve the estimated object velocity.
   In \cite{chadwickDistantVehicleDetection2019}, Chadwick \etal project radar detections to the image plane and use them to 
   boost the object detection accuracy for distant objects. In 
   \cite{nabatiRadarCameraSensorFusion2020} authors use radar detections to 
   generate 3D object proposals first, then project them to the image plane 
   to perform joint 2D object detection and depth estimation. CRF-Net 
   \cite{nobisDeepLearningbasedRadar2019} also projects radar detections to the 
   image plane, but represents them as vertical lines where the pixel values 
   correspond to the depth of each detection point. The image data is then augmented 
   with the radar information and used in a convolutional network to perform 2D object 
   detection.
   
   \begin{figure*}[ht!]
      \includegraphics[width=0.99\textwidth]{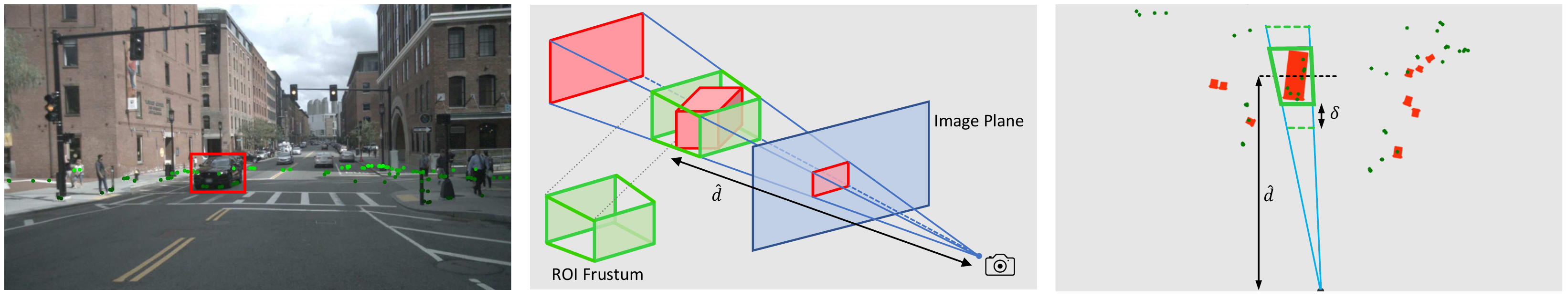}
      \caption{Frustum association. An object detected using the image 
      features (left), generating the ROI frustum based on object's 3D bounding
      box (middle), and the BEV of the ROI frustum showing radar detections 
      inside the frustum (right). $\delta$ is used to increase the frustum size 
      in the testing phase. $\hat{d}$ is the ground truth depth in the training 
      phase and the estimated object depth in the testing phase.}
      \label{fig:frustum}
   \end{figure*}
   
   \begin{figure}[ht!]
      \includegraphics[width=0.48\textwidth]{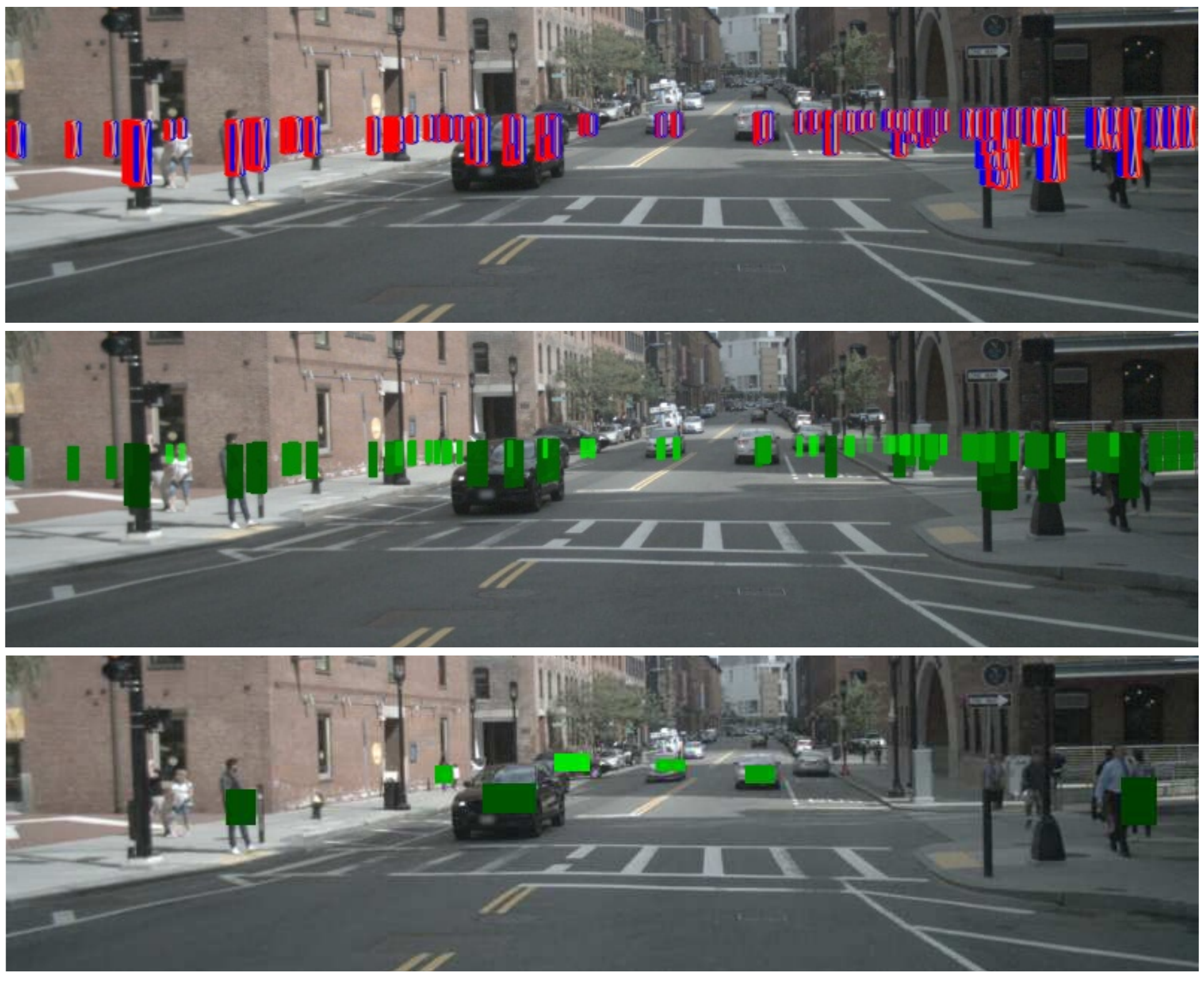}
      \caption{Expanding radar points to 3D pillars (top image). 
           Directly mapping the pillars to the image and replacing with 
           radar depth information results in poor association with objects' 
           center and many overlapping depth values (middle image). 
           Frustum association accurately maps the radar detections
           to the center of objects and minimizes overlapping (bottom image).
           Radar detections are only associated to objects with a valid ground 
           truth or detection box, and only if all or part of the radar detection 
           pillar is inside the box. Frustum association also prevents associating 
           radar detections caused by background objects such as buildings to 
           foreground objects, as seen in the case of pedestrians on the right 
           hand side of the image.}
      \label{fig:pillars}
   \end{figure}

   \section{Preliminary}
   \subsection{Radar Point Cloud}
   Radars are active sensors that transmit radio waves to sense the environment 
   and measure the reflected waves to determine the location and velocity of 
   objects. Automotive radars usually report the detected objects as 2D points in 
   BEV, providing the azimuth angle and radial distance to the object. 
   For every detection, the radar also reports the instantaneous velocity of the 
   object in the radial direction. This radial velocity does not 
   necessarily match the object's actual velocity vector in it's direction of movement.
   Fig. \ref{fig:radarVel} illustrates the difference between the radial as reported
   by the radar, and actual velocity of the object in the vehicle's coordinate system.
   
   We represent each radar detection as a 3D point in the egocentric coordinate
   system, and parameterize it as $P=(x,y,z,v_x,v_y$) where $(x,y,z)$ is the
   position and $(v_x, v_y)$ is the reported radial velocity of the object in the 
   $x$ and $y$ directions. The radial velocity is compensated by the ego vehicle's 
   motion. For every scene, we aggregate 3 sweeps of the radar point cloud 
   (detections within the past 0.25 seconds). The nuScenes dataset provides the 
   calibration parameters needed for mapping the radar point clouds from the 
   radar coordinates system to the egocentric and camera coordinate systems.
   
   \subsection{CenterNet}
   CenterNet \cite{zhou2019objects} represents the state-of-the-art in 3D object detection using single camera. It takes an image $I \in \mathbb{R}^{W \times H \times 3}$ as input and 
   generates a keypoint heatmap $\hat{Y} \in [0,1]^{\frac{W}{R} \times \frac{H}{R} \times C}$
   as output where $W$ and $H$ are the image width and height, $R$ is the 
   downsampling ratio and $C$ is the number of object categories. A prediction of 
   $\hat{Y}_{x,y,c} = 1$ as the output indicates a detected object of class $c$ 
   centered at position $(x,y)$ on the image. 
   The ground-truth heatmap $Y \in [0,1]^{\frac{W}{R} \times \frac{H}{R} \times C}$ 
   is generated from the ground-truth 2D bounding boxes using a Gaussian kernel. 
   For each bounding box center point $p_i \in \mathcal{R}^2$ of class $c$ in the 
   image, a Gaussian heatmap is generated on $Y_{:,:,c}$. The final value of 
   $Y$ for class $c$ at position $q \in \mathcal{R}^2$ is defined as 
   \cite{zhou2019objects}:\useshortskip
   \begin{equation}
       \label{eq1}
       Y_{qc} = \max_{i} \exp(-\frac{(p_{i}-q)^2}{2\sigma_{i}^2})
   \end{equation}
   where $\sigma_i$ is a size-adaptive standard deviation, controlling the size 
   of the heatmap for every object based on its size.
   A fully convolutional encode-decoder network is used to predict $\hat{Y}$.
   
   To generate 3D bounding boxes, separate network heads are used to regress 
   object's depth, dimensions and orientation directly from the detected center 
   points. Depth is calculated as an additional output channel 
   $\hat{D}\in [0,1]^{\frac{W}{R} \times \frac{H}{R}}$ after applying the inverse 
   sigmoidal transformation used in Eigen \etal \cite{eigen2014depth} to the 
   original depth domain. The object dimensions are directly regressed to their 
   absolute values in meter as three output channels 
   $\hat{\Gamma}\in [0,1]^{\frac{W}{R} \times \frac{H}{R}\times 3}$. Orientation is 
   encoded as two bins with 4 scalars in each bin, following the orientation 
   representation in Mousavian \etal \cite{mousavian20173d}. For each center point, a 
   local offset is also predicted to compensate for the discretization error 
   caused by the output strides in the backbone network \cite{zhou2019objects}.
   
   Given the annotated objects ${p_0,p_1,...}$ in an image, the training objective
   is defined as below based on the focal loss \cite{linFocalLossDense2018}:
   
   \small
   \begin{equation*}
       L_{k} = \frac{1}{N} \sum_{xyc}
       \begin{cases}
           (1 - \hat{Y}_{xyc})^{\alpha} 
           \log(\hat{Y}_{xyc}) & \! Y_{xyc}=1\vspace{2mm}\\
           (1-Y_{xyc})^{\beta} 
           (\hat{Y}_{xyc})^{\alpha}\log(1-\hat{Y}_{xyc})
           & \!\text{otherwise}
       \end{cases},
   \end{equation*}
   \normalsize
   where $N$ is the number of objects, 
   $Y \in [0,1]^{\frac{W}{R} \times \frac{H}{R} \times C}$ is the annotated 
   objects' ground-truth heatmap and $\alpha$ and $\beta$ are focal loss 
   hyperparameters.

   \section{\method{}}
   
   In this section we present our approach to radar and camera sensor fusion for 
   3D object detection. The overall \method{} architecture is
   shown in Fig. \ref{fig:netArch}. We adopt \cite{zhou2019objects} 
   as our center-based object detection network to detect
   objects' center points on the image plane, and regress to other object 
   properties such as 3D location, orientation and dimensions. 
   We propose a middle-fusion mechanism 
   that associates radar detections to their corresponding object's center 
   point and exploits both radar and image features to improve the preliminary
   detections by re-estimating their depth, velocity, rotation and attributes.
   
   The key in our fusion mechanism is accurate association of radar detections 
   to objects. The center point object detection network 
   generates a heat map for every object category in the image. 
   The peaks in the heat map represent possible center points for objects, 
   and the image features at those locations are used to estimate other object 
   properties. To exploit the radar information in this setting, radar-based features
   need to be mapped to the center of their corresponding object on the image,
   which requires an accurate association between the radar detections and objects
   in the scene.
   
   \subsection{Center Point Detection}
   We adopt the CenterNet \cite{zhou2019objects} detection network for generating
   preliminary detections on the image. The image features are first extracted 
   using a fully convolutional encoder-decoder backbone network. We follow 
   CenterNet \cite{zhou2019objects} and use a modified version of the Deep Layer
   Aggregation (DLA) network \cite{yuDeepLayerAggregation2018} as the backbone.
   The extracted image features are then used
   to predict object center points on the image, as well as
   the object 2D size (width and height), center offset, 3D dimensions, depth 
   and rotation. These values are predicted by the primary regression heads as shown 
   in Fig. \ref{fig:netArch}. Each primary regression head consists of a $3\times3$ 
   convolution layer with 256 channels and a $1\times1$ convolutional layer to 
   generate the desired output. This provides an accurate 2D bounding box as well 
   as a preliminary 3D bounding box for every detected object in the scene.

   \subsection{Radar Association}
   
   The center point detection network only uses the image features at the center
   of each object to regress to all other object properties.
   To fully exploit radar data in this process, we first need to associate the radar 
   detections to their corresponding object on the image plane. To accomplish 
   this, a na\"{i}ve approach would be mapping each radar detection point to the 
   image plane and associating it to an object if the point is mapped inside 
   the 2D bounding box of that object. This is not a very robust solution, as 
   there is not a one-to-one mapping between radar detections and objects
   in the image; Many objects in the scene generate multiple radar 
   detections, and there are also radar detections that do not correspond to 
   any object.
   Additionally, because the $z$ dimension of the radar detection is not accurate
   (or does not exist at all), the mapped radar detection might end up outside 
   the 2D bounding box of its corresponding object. Finally, radar detections
   obtained from occluded objects would map to the same general area in the 
   image, which makes differentiating them in the 2D image plane difficult, 
   if possible at all. 
   
   \textbf{Frustum Association Mechanism:}
   We develop a frustum association method that uses the object's 2D bounding 
   box as well as its estimated depth and size to create a 3D Region of Interest (RoI)
   frustum for the object. Having an accurate 2D bounding box for an object, we create 
   a frustum for that object as shown in Fig. \ref{fig:frustum}. This significantly
   narrows down the radar detections that need to be checked for association, as any
   point outside this frustum can be ignored. We then use the estimated object depth,
   dimension and rotation to create a RoI around the object, to further filter out 
   radar detections that are not associated with this object. If there are multiple 
   radar detections inside this RoI, we take the closest point as the radar detection 
   corresponding to this object. 
   
   In the training phase, we use the object's 3D ground truth bounding box
   to create a tight RoI frustum and associate radar detections to the object. 
   In the test phase, the RoI frustum is calculated 
   using the object's estimated 3D bounding box as explained before. In this case, 
   we use a parameter $\delta$ to control the size of the RoI frustum
   as shown in Fig. \ref{fig:frustum}.
   This is to account for inaccuracy in the estimated depth values, as the depth 
   of the object at this stage is solely determined using the image-based features. 
   Enlarging the frustum using this parameter increases the chance
   of including the corresponding radar detections inside the frustum even if 
   the estimated depth is slightly off. The value of $\delta$ should be carefully 
   selected, as a large RoI frustum can include radar detections of nearby objects.
   
   The RoI frustum approach makes associating
   overlapping objects effortless, as objects are separated in the 3D space 
   and would have separate RoI frustums. It also eliminates the multi-detection 
   association problem, as only the closest radar detection inside the RoI frustum
   is associated to the object. It does not, however, help with the inaccurate 
   $z$ dimension problem, as radar detections might be outside the 
   ROI frustum of their corresponding object due to their inaccurate height 
   information. 
   
   \textbf{Pillar Expansion:}
   To address the inaccurate height information problem, we introduce a radar 
   point cloud preprocessing 
   step called pillar expansion, where each radar point is expanded to 
   a fixed-size pillar, as illustrated in Fig. \ref{fig:pillars}.
   Pillars create a better representation for the physical objects detected
   by the radar, as these detections are now associated with a dimension in 
   the 3D space. Having this new representation, we simply consider a radar detection to be inside a frustum 
   if all or part of its corresponding pillar is inside the frustum, as shown 
   in Fig. \ref{fig:netArch}.

   \subsection{Radar Feature Extraction}
   After associating radar detections to their corresponding objects, we use the 
   depth and velocity of the radar detections to create complementary features for
   the image. Particularly, for every radar detection associated to an object, 
   we generate three heat map channels centered at and inside the object's 2D bounding 
   box, as shown in Fig. \ref{fig:pillars}. 
   The width and height of the heatmaps are proportional to the object's 2D bounding
   box, and are controlled by a parameter $\alpha$.
   The heatmap values are the normalized object depth ($d$) and also the $x$ and $y$ 
   components of the radial velocity ($v_x$ and $v_y$) in the egocentric coordinate system:
   
   \begin{equation*}
       F_{x,y,i}^{j} = \frac{1}{M_i}
       \begin{cases}
           f_i & \! |x-c_{x}^{j}|\leq \alpha w^j \hspace{8pt} \text{and}\\
           &|y-c_{y}^{i}|\leq \alpha h^j \\
           0 & \!\text{otherwise}
       \end{cases},
   \end{equation*}

   where $i \in {1,2,3}$ is the feature map channel, $M_i$ is a normalizing factor,
   $f_i$ is the feature value ($d$, $v_x$ or $v_y$), $c_{x}^{j}$ and $c_{y}^{j}$ are 
   the $x$ and $y$ coordinates of the $j$th object's center point on the image and 
   $w^j$ and $h^j$ are the width and height of the $j$th object's 2D bounding box.
   If two objects have overlapping heatmap areas, the one with a smaller depth value
   dominates, as only the closest object is fully visible in the image.
   
   The generated heat maps are then concatenated to the image features as extra channels. 
   These features are used as inputs to the secondary regression heads to recalculate 
   the object's depth and rotation, as well as velocity 
   and attributes. The velocity regression head estimates
   the $x$ and $y$ components of the object's actual velocity in the vehicle coordinate
   system. The attribute regression head estimates different attributes for different 
   object classes, such as moving or parked for the Car class and standing or sitting
   for the Pedestrian class. The secondary regression heads consist of three convolutional 
   layers with $3\times3$ kernels followed by a $1\times1$ convolutional layer to generate
   the desired output. The extra convolutional layers compared to the primary regression 
   heads help with learning higher level features from the radar feature maps. The last 
   step is decoding the regression head results into 3D bounding boxes. The box decoder
   block uses the estimated depth, velocity, rotation, and attributes from the secondary
   regression heads, and takes the other object properties from the primary heads.

   \section{Implementation Details}
   We use the pre-trained CenterNet \cite{zhou2019objects} network with the 
   DLA \cite{yuDeepLayerAggregation2018} backbone as our object detection network. 
   DLA uses iterative deep aggregation layers to increase the resolution 
   of feature maps. CenterNet compares its performance using different backbone 
   architectures, with the Hourglass network \cite{newellHourglass2016a} performing
   better than others. We choose to use the DLA network as it takes significantly less
   time to train while providing a reasonable performance.
   
   We directly use the released CenterNet model that is trained for 140 epochs
   on the nuScenes dataset. This model by default does not 
   provide velocity and attribute predictions. We train the velocity and attribute 
   heads for 30 epochs, and use the resulting model as our baseline. 
   The secondary regression heads in our network are added on top of the CenterNet 
   backbone network, and are trained using
   the image and radar features for an additional 60 epochs with a batch size 
   of 26 on two Nvidia P5000 GPUs. 
   
   During both training and testing, we reduce the image resolution from the original
   1600$\times$900 pixels to 800$\times$450 pixels. Data augmentation is used 
   during training, with random right-left flipping (with a probability of 0.5) 
   and random shifting (from 0 to 20 percent of image size). The same augmentations
   are also applied to the radar point cloud in reference to the camera coordinate 
   system. We do not apply any scaling augmentation as it changes the 3D
   measurements. At testing time, we only use flip test augmentation where 
   an image and its flipped version are fed into the network and 
   the average of the network outputs is used for decoding the 3D bounding 
   boxes.
   We do not use the multi-scale test augmentation as used by 
   CenterNet. The pillar size is set to $[0.2, 0.2, 1.5]$  meters in the 
   $[x,y,z]$ directions and $\delta$ is 
   set to increase the length of the RoI frustum by 20\% in the radial direction 
   at test time.
   
   We use the L1 loss for most of the regression heads, with the exception of 
   the center point heat map head which uses the focal loss and the attributes
   regression head that uses the Binary Cross Entropy (BCE) loss.

   \newcolumntype{g}{>{\columncolor{Gray}}c@{\hskip0.2cm}}
   \newcolumntype{C}{>{\hskip0.1cm}c@{\hskip0.1cm}}
   \newcolumntype{D}{>{\hskip0.05cm}c@{\hskip0.0cm}}

   \begin{table*}[ht!]
       \centering
       \caption{Performance comparison for 3D object detection on nuScenes dataset.
           mATE, mASE, mAOE, mAVE and mAAE stand for average translation, scale, orientation, 
           velocity and attribute errors respectively. $\uparrow$ indicates that higher is
           better and $\downarrow$ indicates that lower is better. "C", "R" and 
           "L" specify camera, radar and LIDAR modalities respectively.}
        \begin{tabular}{l C C C C@{\hskip 0.3cm} g C C C C C c}

           \hline
           & & \multicolumn{3}{c}{Modality} & & & \multicolumn{5}{c}{Error $\downarrow$} \\
           \cline{3-5} \cline{8-12}
           Method & Dataset & C & R & L & NDS $\uparrow$ & mAP $\uparrow$ & mATE & mASE & mAOE & mAVE & mAAE \\ 
           \hline
           InfoFocus \cite{wang2020infofocus} & test & & & \checkmark & 0.395 & \textbf{0.395} & \textbf{0.363} & 0.265 & 1.132 & 1.000 & 0.395 \\
           OFT \cite{roddick2018orthographic} & test & \checkmark & & & 0.212 & 0.126 & 0.820 & 0.360 & 0.850 & 1.730 & 0.480 \\
           MonoDIS \cite{simonelli2019a} & test & \checkmark & & & 0.384 & 0.304 & 0.738 & 0.263 & 0.546 & 1.533 & 0.134 \\
           CenterNet (HGLS) \cite{zhou2019objects} & test & \checkmark & & & 0.400 & 0.338 & 0.658 & \textbf{0.255} & 0.629 & 1.629 & 0.142 \\
           Ours (DLA) & test & \checkmark & \checkmark & & \textbf{0.449} & 0.326 & 0.631 & 0.261 & \textbf{0.516} & \textbf{0.614} & \textbf{0.115} \\
           \hline
           \hline
           CenterNet (DLA) \cite{zhou2019objects} & val & \checkmark & & & \cellcolor{Gray} 0.328 & 0.306 & 0.716 & 0.264 & 0.609 & 1.426 & 0.658 \\
           Ours (DLA) & val & \checkmark & \checkmark & & \cellcolor{Gray} \textbf{0.453} & \textbf{0.332} & \textbf{0.649} & \textbf{0.263} & \textbf{0.535} & \textbf{0.540} & \textbf{0.142} \\
       \end{tabular}
       \label{res:score}
   \end{table*}
   
   \begin{table*}[ht!]
       \centering
       \caption{Per-class performance comparison for 3D object detection on nuScenes dataset.}
        \begin{tabular}{@{\extracolsep{4pt}}l@{\hskip0.1cm} D D D D@{\hskip 0.2cm} D D D D D D D D D D}
           \hline
           &  & \multicolumn{3}{c}{Modality} & \multicolumn{10}{c}{mAP $\uparrow$} \\
           \cline{3-5} \cline{6-15}
           Method & Dataset & C & R & L & Car & Truck & Bus & Trailer & Const. & Pedest. & Motor. & Bicycle & Traff. & Barrier\\
           \hline
           InfoFocus \cite{wang2020infofocus} & test & & & \checkmark & \textbf{0.779} & \textbf{0.314} & \textbf{0.448} & \textbf{0.373} & \textbf{0.107} & \textbf{0.634} & 0.290 & 0.061 & 0.465 & 0.478 \\
           MonoDIS \cite{simonelli2019a} & test & \checkmark & & & 0.478 & 0.220 & 0.188 & 0.176 & 0.074 & 0.370 & 0.290 & \textbf{0.245} & 0.487 & 0.511 \\
           CenterNet (HGLS) \cite{zhou2019objects} & test & \checkmark & & & 0.536 & 0.270 & 0.248 & 0.251 & 0.086 & 0.375 & 0.291 & 0.207 & \textbf{0.583} & \textbf{0.533} \\
           Ours (DLA) & test & \checkmark & \checkmark & & 0.509 & 0.258 & 0.234 & 0.235 & 0.077 & 0.370 & \textbf{0.314} & 0.201 & 0.575 & 0.484 \\
           \hline 
           \hline
           CenterNet (DLA) \cite{zhou2019objects} & val & \checkmark & &  & 0.484 & 0.231 & 0.340 & 0.131 & 0.035 & 0.377 & 0.249 & \textbf{0.234} & 0.550 & 0.456 \\
           Ours (DLA) & val & \checkmark & \checkmark & &  \textbf{0.524} & \textbf{0.265} & \textbf{0.362} & \textbf{0.154} & \textbf{0.055} & \textbf{0.389} & \textbf{0.305} & 0.229 & \textbf{0.563} & \textbf{0.470} \\
       \end{tabular}
       \label{res:class}
   \end{table*}

   \section{Results}
   We compare our radar and camera fusion network with the state-of-the-art camera-based
   models on the nuScenes benchmark, and also a LIDAR based method. 
   Table \ref{res:score} shows the results
   on both test and validation splits of the nuScenes dataset. We compare 
   with OFT \cite{roddick2018orthographic}, MonoDIS \cite{simonelli2019a} and
   CenterNet \cite{zhou2019objects}
   which are camera-based 3D object detection networks, as well as InfoFocus
   \cite{wang2020infofocus} which is a LIDAR-based method. As seen in 
   Table \ref{res:score}, CenterFusion outperforms 
   all other methods in the nuScenes NDS score, which is a weighted sum of the 
   mAP and the error metrics.
   On the test dataset, CenterFusion 
   shows a 12.25\% and 16.9\% relative increase in the NDS score compared to CenterNet and 
   MonoDIS respectively. The LIDAR-based method InfoFocus shows a better 
   performance in the mAP score compared to other methods, but is significantly 
   outperformed by CenterFusion 
   in the orientation, velocity and attribute error metrics. While CenterNet with the 
   Hourglass \cite{newellHourglass2016a} backbone network results in a better
   mAP score compared to CenterFusion (1.2\% difference) on the test split, 
   the results on the 
   validation split show that CenterFusion outperforms CenterNet by 2.6\% when 
   both networks use the same DLA \cite{yuDeepLayerAggregation2018} backbone.
   The validation set results also show CenterFusion 
   improving CenterNet in all the other metrics. CenterFusion shows an absolute 
   gain of 38.1\% and 62.1\% relative increase in the NDS 
   and velocity error metrics compared to CenterNet, which demonstrates the 
   effectiveness of using radar features.
   
   Table \ref{res:class} compares the per-class mAP results for both 
   test and validation splits. While CenterNet with an Hourglass backbone has a 
   higher mAP than CenterFusion for most classes in the test set, it is 
   outperformed by CenterFusion on the validation set where
   the DLA backbone is used for both methods. The most improved classes 
   on the validation set are the motorcycle and car with 5.6\% and 4.0\% absolute 
   mAP increase respectively.
   
\begin{figure*}[ht!]
       \minipage{0.24\textwidth}
           \includegraphics[width=\linewidth]{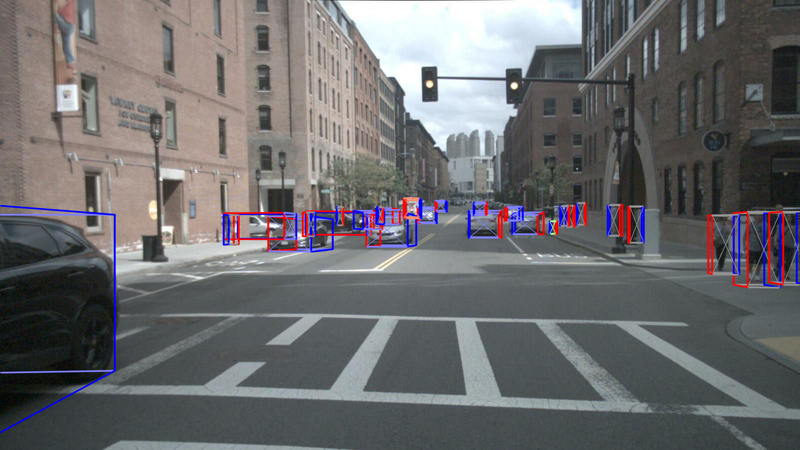}
       \endminipage \hspace{1pt}
       \minipage{0.24\textwidth}
           \includegraphics[width=\linewidth]{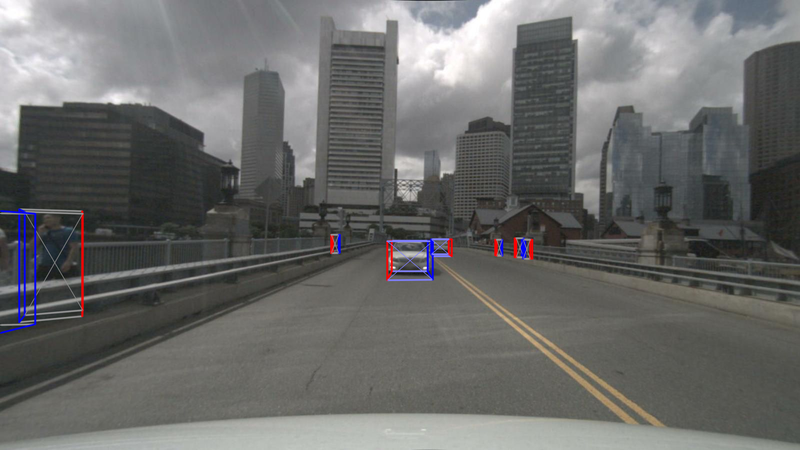}
       \endminipage \hspace{1pt}
       \minipage{0.24\textwidth}
           \includegraphics[width=\linewidth]{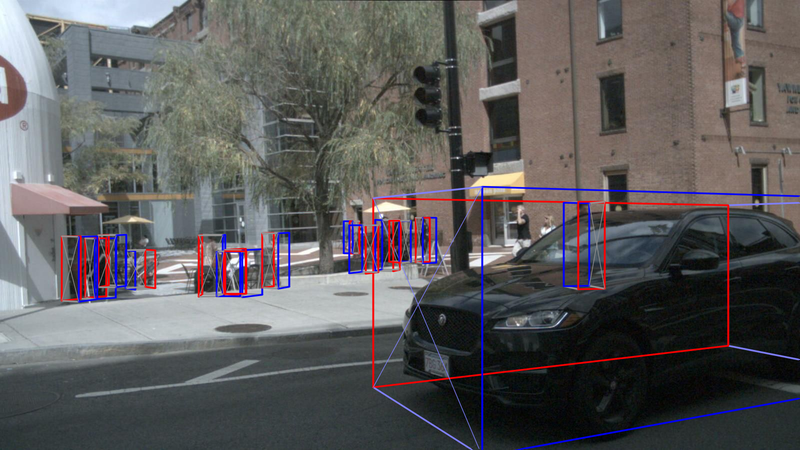}
       \endminipage \hspace{1pt}
       \minipage{0.24\textwidth}
           \includegraphics[width=\linewidth]{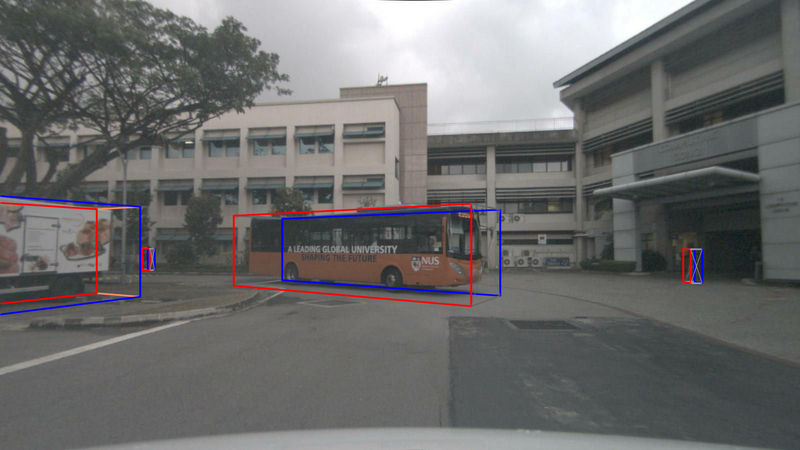}
       \endminipage\\
       \minipage{0.24\textwidth}
           \includegraphics[width=\linewidth]{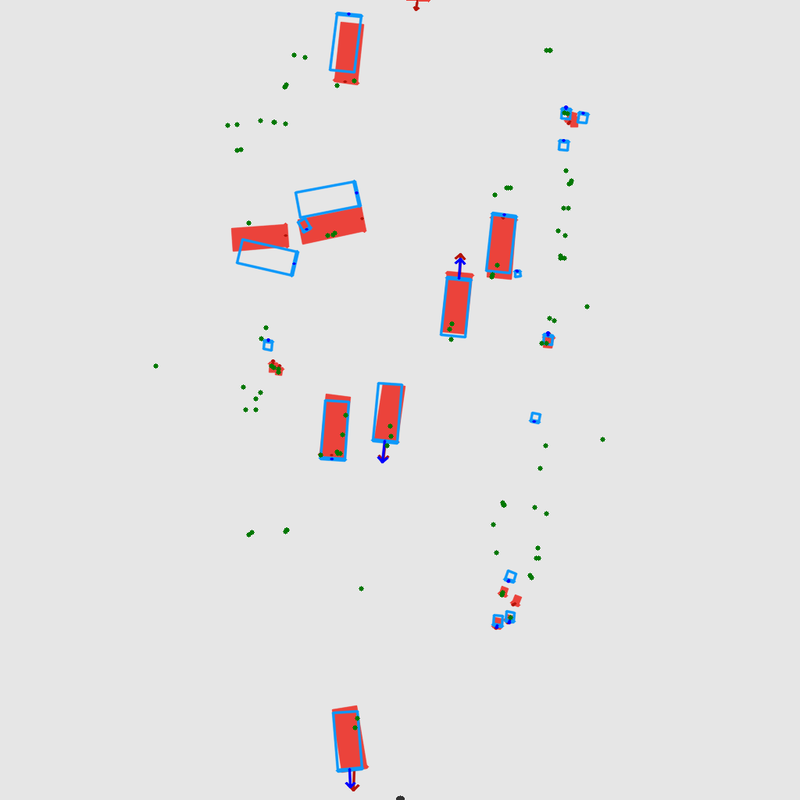}
       \endminipage \hspace{1pt}
       \minipage{0.24\textwidth}
           \includegraphics[width=\linewidth]{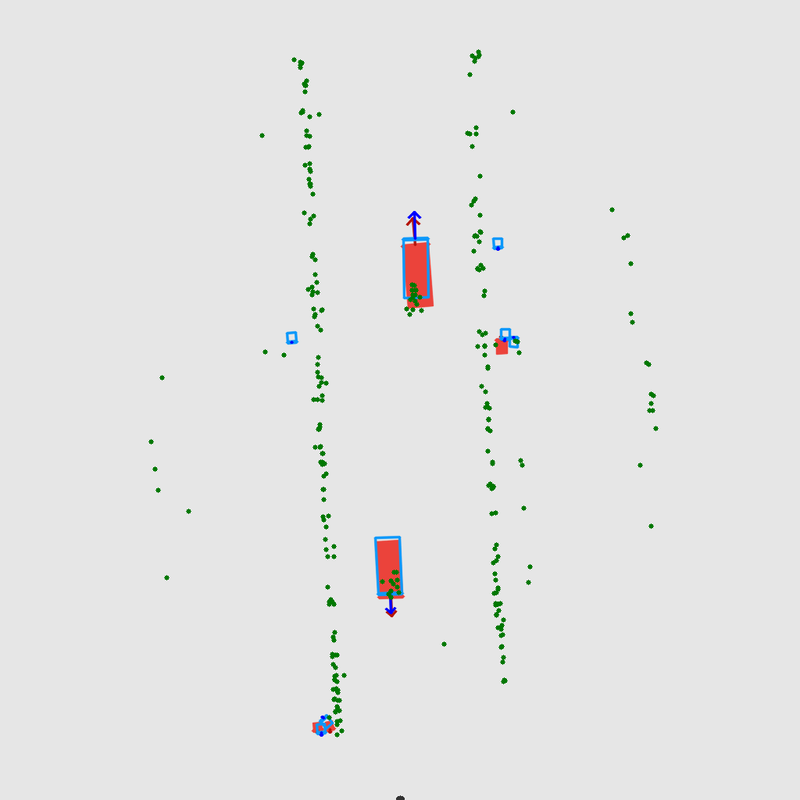}
       \endminipage \hspace{1pt}
       \minipage{0.24\textwidth}
           \includegraphics[width=\linewidth]{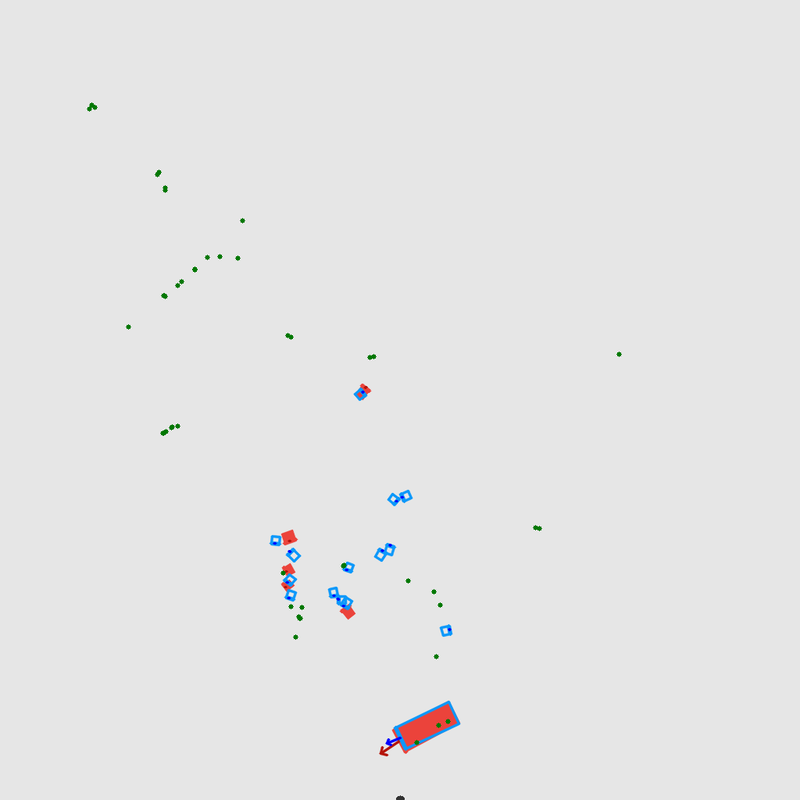}
       \endminipage \hspace{1pt}
       \minipage{0.24\textwidth}
           \includegraphics[width=\linewidth]{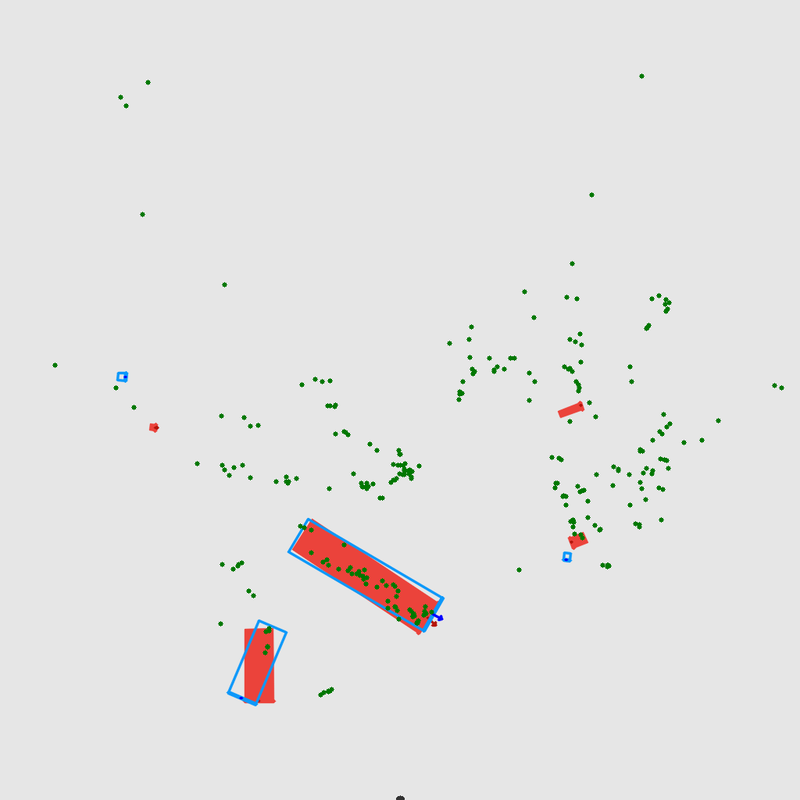}
       \endminipage\\[3pt]
       \minipage{0.24\textwidth}
           \includegraphics[width=\linewidth]{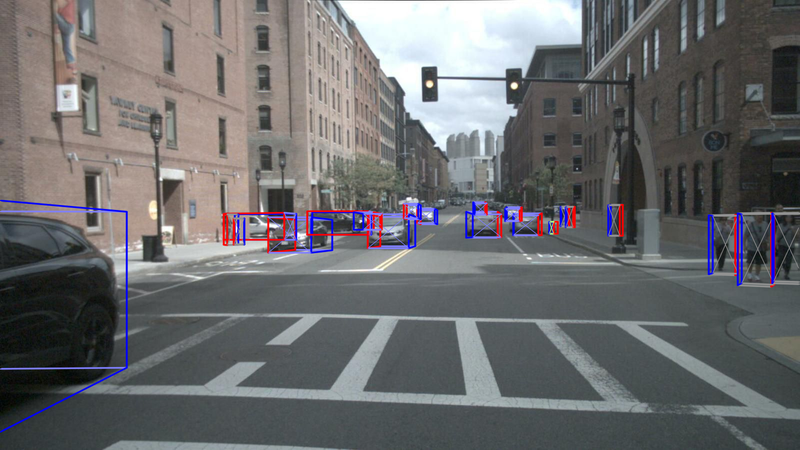}
       \endminipage \hspace{1pt}
       \minipage{0.24\textwidth}
           \includegraphics[width=\linewidth]{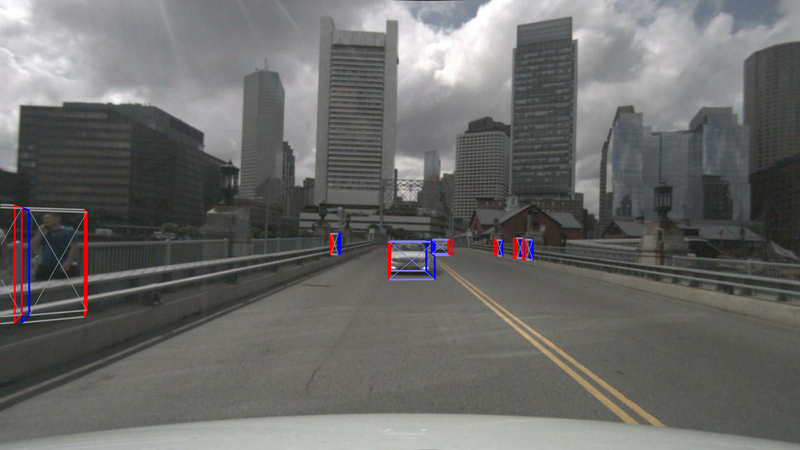}
       \endminipage \hspace{1pt}
       \minipage{0.24\textwidth}
           \includegraphics[width=\linewidth]{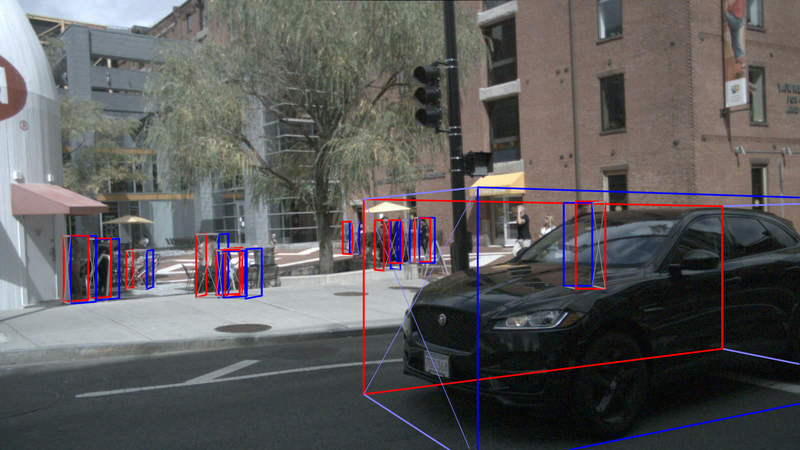}
       \endminipage \hspace{1pt}
       \minipage{0.24\textwidth}
           \includegraphics[width=\linewidth]{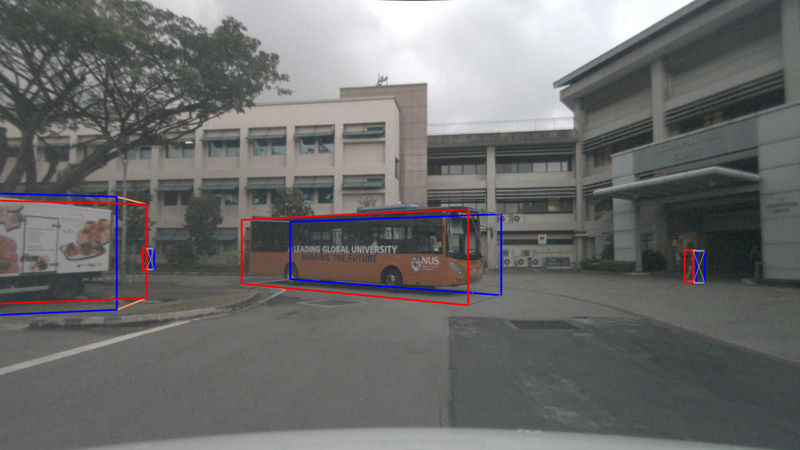}
       \endminipage\\
       \minipage{0.24\textwidth}
           \includegraphics[width=\linewidth]{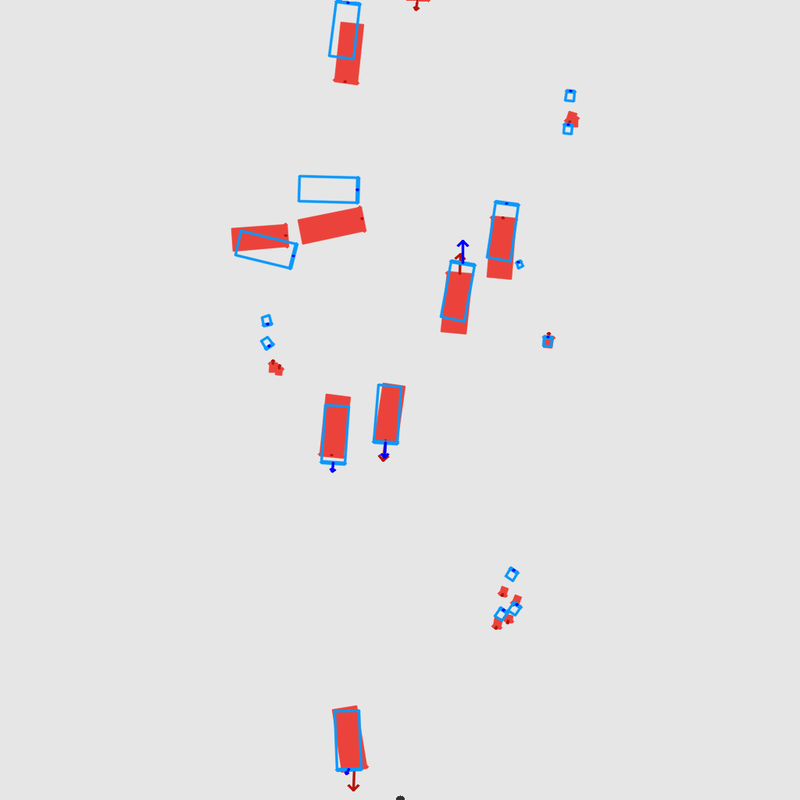}
       \endminipage \hspace{1pt}
       \minipage{0.24\textwidth}
           \includegraphics[width=\linewidth]{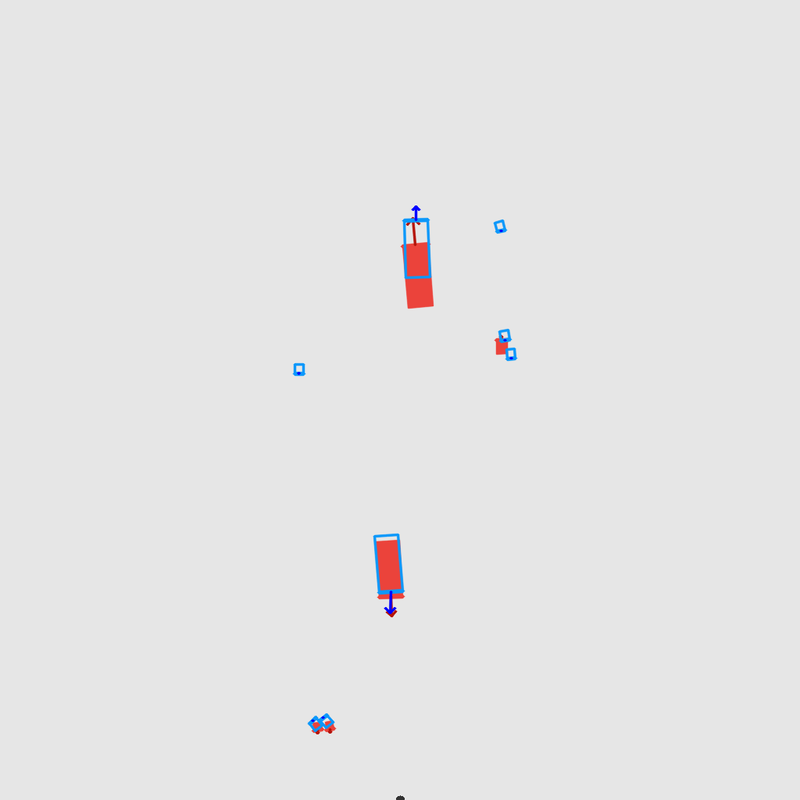}
       \endminipage \hspace{1pt}
       \minipage{0.24\textwidth}
           \includegraphics[width=\linewidth]{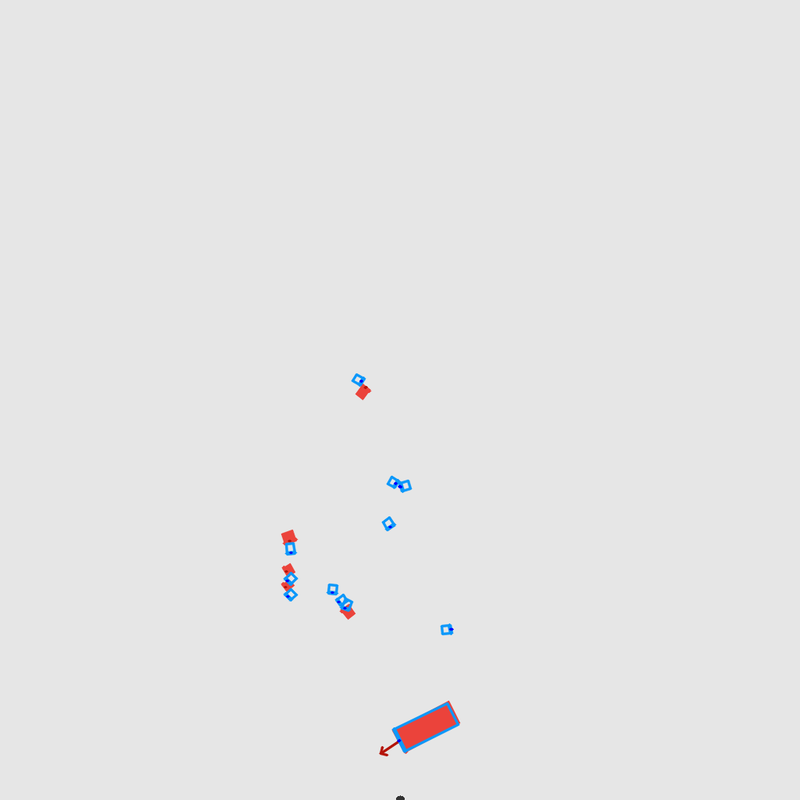}
       \endminipage \hspace{1pt}
       \minipage{0.24\textwidth}
           \includegraphics[width=\linewidth]{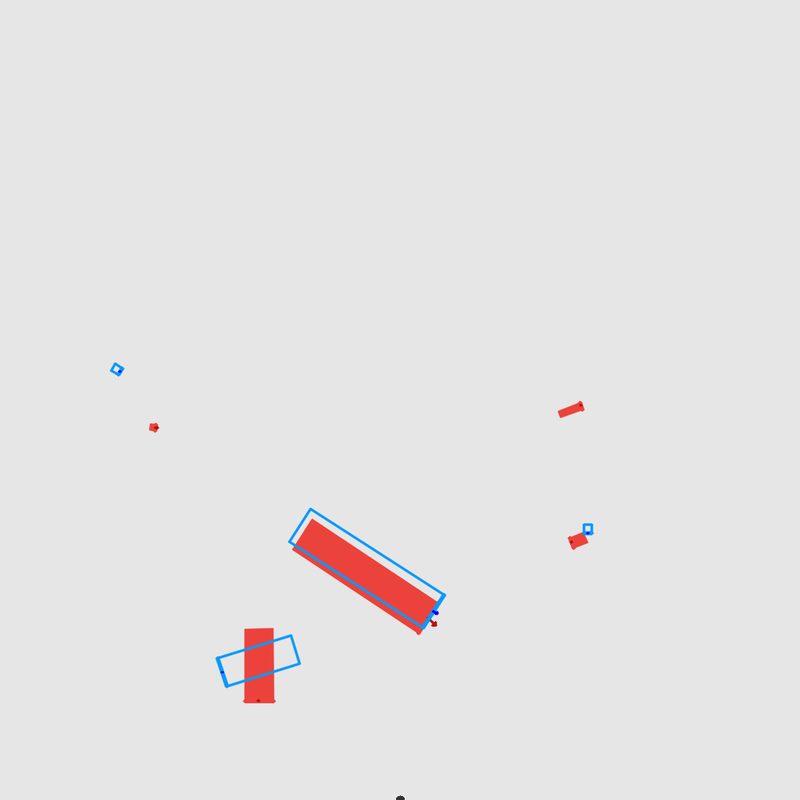}
       \endminipage\\
      \caption{Qualitative results from CenterFusion (row 1 \& 2) and CenterNet
           (row 3 \& 4) in camera view and BEV. In the BEV plots, 
           detection boxes are shown in {\color{cyan} cyan} and ground truth boxes
           in {\color{red} red}. The radar point cloud is shown in 
           {\color{ForestGreen} green}. Red and blue arrows on objects show the ground 
           truth and predicted velocity vectors respectively.}
      \label{fig:res}
   \end{figure*}

   Fig. \ref{fig:res} demonstrates the 3D object detection results in both camera
   and BEV. It shows the detection results from CenterFusion (row 1 \& 2) and 
   CenterNet (row 3 \& 4) for 4 different scenes. The radar point clouds are 
   also shown in the CenterFusion BEV results.
   Compared to CenterNet, the results from CenterFusion show 
   a better fit for 3D boxes in most cases, especially objects at a larger distance, 
   such as the far vehicle in the second scene. 
   Additionally, the velocity vectors estimated by CenterFusion show a 
   significant improvement compared to the CenterNet results, as seen 
   in the second and third scenes.

   \begin{table*}[ht!]
    \centering
    \caption{Overall ablation study on nuScenes validation set. Improvement percentages 
    in each row are relative to the baseline method. (PE: Pillar Expansion, FA: Frustum Association, FT: Flip Test)}
    \begin{tabular}{lc@{\hskip0.1cm}c@{\hskip0.2cm}c@{\hskip0.2cm}c@{\hskip0.2cm}c@{\hskip0.4cm}c@{\hskip0.2cm}c@{\hskip0.2cm}c@{\hskip0.2cm}c@{\hskip0.2cm}c@{\hskip0.2cm}c@{\hskip0.2cm}c}
        \hline
        Method & Cam & Rad & PE & FA & FT & NDS $\uparrow$ & mAP $\uparrow$ & mATE $\downarrow$ & mASE $\downarrow$ & mAOE $\downarrow$ & mAVE $\downarrow$ & mAAE $\downarrow$ \\ 
        \hline
        Baseline & \checkmark & - & - & - & - & 0.328 & 0.306 & 0.716 & 0.264 & 0.609 & 1.426 & 0.658 \\
        \hline
        Ours & \checkmark & \checkmark & \checkmark & - & - & +15.4\% & +1.0\%& -2.0\% & +1.1\%&  -4.4\%& -13.1\%& -68.6\% \\
        Ours & \checkmark & \checkmark & - & \checkmark & - & +25.9\% & +2.0\%& -2.8\% & +1.0\%&  -7.4\%& -48.1\%& -75.9\% \\
        Ours & \checkmark & \checkmark & \checkmark & \checkmark & - & +34.5\% & +4.3\%& -5.3\% & +1.1\%&  -10.0\%& -61.9\%& -78.0\% \\
        Ours & \checkmark & \checkmark & \checkmark & \checkmark & \checkmark & \textbf{+37.8\%} & \textbf{+8.4\%} & \textbf{-9.4\%} & \textbf{-0.5\%} &  \textbf{-11.6\%} & \textbf{-62.0\%} & \textbf{-78.3\%} \\
    \end{tabular}
    \label{table:abl-score}
\end{table*}

\begin{table*}[ht!]
    \centering
    \caption{Class-based ablation study results on nuScenes validation set.}
    \begin{tabular}{lc@{\hskip0.1cm}c@{\hskip0.2cm}c@{\hskip0.2cm}c@{\hskip0.2cm}c@{\hskip0.4cm}c@{\hskip0.2cm}c@{\hskip0.2cm}c@{\hskip0.2cm}c@{\hskip0.2cm}c@{\hskip0.1cm}c@{\hskip0.1cm}c@{\hskip0.1cm}c@{\hskip0.1cm}c@{\hskip0.1cm}c}
        \hline
        Method & Cam & Rad & PE & FA & FT & Car & Truck & Bus & Trailer & Const. & Pedest. & Motor. & Bicycle & Traff. & Barrier\\
        \hline
        Baseline & \checkmark & - & - & - & - & 48.4 & 23.1 & 34.0 & 13.1 & 3.5 & 37.7 & 24.9 & 23.4 & 55.0 & 45.6 \\
        \hline
        Ours & \checkmark & \checkmark & \checkmark & - & - & +0.6 & +0.7 & -2.1 & +0.9 & +0.6 & +0.9 & +1.9 & -2.5 & +0.1 & +0.8 \\
        Ours & \checkmark & \checkmark & - & \checkmark & - & +1.0 & +1.0 & -2.1 & +0.9 & +0.9 & 0.0 & +2.1 & -1.9 & +0.2 & +0.8 \\
        Ours & \checkmark & \checkmark & \checkmark & \checkmark & - & +2.8 & +2.1 & -1.2 & +1.4 & +1.1 & +0.1 & +3.8 & -1.1 & +0.4 & +0.8 \\
        Ours & \checkmark & \checkmark & \checkmark & \checkmark & \checkmark & \textbf{+4.1} & \textbf{+3.4} & \textbf{+2.7} & \textbf{+1.8} &  \textbf{+1.8} & \textbf{+1.2} & \textbf{+5.5} & -0.7 & \textbf{+1.3} & \textbf{+1.5} \\
    \end{tabular}
    \label{table:abl-class}
\end{table*}

   \section{Ablation Study}
   We validate the effectiveness of our fusion approach by conducting an ablation study
   on the nuScenes validation set. We use the CenterNet model as our baseline, and 
   study the effectiveness of the pillar expansion, frustum association and flip testing
   on the detection results. Table \ref{table:abl-score} shows the overall detection
   results of the ablation study.
   
   In the first experiment, we only apply pillar expansion to the radar point clouds,
   and map the 3D pillars to the image plane and obtain their equivalent 2D bounding boxes.
   These boxes are then filled with the depth and velocity values of their corresponding 
   radar detections and used as the radar feature maps, as shown in Fig. \ref{fig:pillars}.
   According to Table \ref{table:abl-score}, this simple association method results in 
   a 15.4\% relative improvement on the NDS score and 1.0\% absolute improvement on the 
   mAP compared to the baseline. 
   
   In the next experiment we only use the frustum association method by directly 
   applying it on the radar point clouds without converting them to pillars first.
   This improves the NDS score by 25.9\% relatively and mAP by 2.0\%. Applying 
   both pillar expansion and frustum association results in a relative 35.5\% and 
   absolute 4.3\% improvement on the NDS and mAP scores respectively.
   Flip testing adds another 3.3\% improvement on the NDS 
   score and 3.9\% on the mAP, resulting in a total of 37.8\% and 8.4\% improvement 
   on NDS and mAP compared to the baseline method. 
   
   Table \ref{table:abl-class} shows the per-class contribution of each step on the mAP.
   According to the results, both pillar expansion and 
   frustum association steps have contributed to the improvement of mAP in most 
   object classes. The only class that has not improved from the baseline is the 
   bicycle class, in which the CenterNet mAP score is better than CenterFusion by 
   0.5\%.

   \section{Conclusion}
   In summary, we proposed a new radar and camera fusion algorithm called CenterFusion,
   to exploit radar information for robust 3D object detection. CenterFusion accurately 
   associates radar detections to objects on the image using a frustum-based 
   association method, and creates radar-based feature maps to complement the image 
   features in a middle-fusion approach. Our frustum association method uses 
   preliminary detection results to generate a RoI frustum around objects in 3D space, 
   and maps the radar detection to the center of objects on the image. We also 
   used a pillar expansion method to compensate for the inaccuracy in radar detections' 
   height information, by converting radar points to fixed-size pillars
   in the 3D space. We evaluated our proposed method on the challenging nuScenes 3D 
   detection benchmark, where CenterFusion outperformed the state-of-the-art camera-based 
   object detection methods.

{\small
\bibliographystyle{ieee_fullname}
\bibliography{references}

\begin{thebibliography}{10}\itemsep=-1pt

\bibitem{7795718}
A. {Asvadi}, P. {Girão}, P. {Peixoto}, and U. {Nunes}.
\newblock 3d object tracking using rgb and lidar data.
\newblock In {\em 2016 IEEE 19th International Conference on Intelligent
  Transportation Systems (ITSC)}, pages 1255--1260, 2016.

\bibitem{nuscenes2019}
Holger Caesar, Varun Bankiti, Alex~H. Lang, Sourabh Vora, Venice~Erin Liong,
  Qiang Xu, Anush Krishnan, Yu Pan, Giancarlo Baldan, and Oscar Beijbom.
\newblock nuscenes: A multimodal dataset for autonomous driving.
\newblock {\em arXiv preprint arXiv:1903.11027}, 2019.

\bibitem{chadwickDistantVehicleDetection2019}
Simon Chadwick, Will Maddern, and Paul Newman.
\newblock Distant {{Vehicle Detection Using Radar}} and {{Vision}}.
\newblock {\em arXiv:1901.10951 [cs]}, May 2019.

\bibitem{Chen_2017}
Xiaozhi Chen, Huimin Ma, Ji Wan, Bo Li, and Tian Xia.
\newblock Multi-view 3d object detection network for autonomous driving.
\newblock {\em 2017 IEEE Conference on Computer Vision and Pattern Recognition
  (CVPR)}, Jul 2017.

\bibitem{chenMultiView3DObject2017}
Xiaozhi Chen, Huimin Ma, Ji Wan, Bo Li, and Tian Xia.
\newblock Multi-{{View 3D Object Detection Network}} for {{Autonomous
  Driving}}.
\newblock {\em arXiv:1611.07759 [cs]}, June 2017.

\bibitem{eigen2014depth}
David Eigen, Christian Puhrsch, and Rob Fergus.
\newblock Depth map prediction from a single image using a multi-scale deep
  network.
\newblock In {\em Advances in neural information processing systems}, pages
  2366--2374, 2014.

\bibitem{fang2019camera}
Yongkun Fang, Huijing Zhao, Hongbin Zha, Xijun Zhao, and Wen Yao.
\newblock Camera and lidar fusion for on-road vehicle tracking with
  reinforcement learning.
\newblock In {\em 2019 IEEE Intelligent Vehicles Symposium (IV)}, pages
  1723--1730. IEEE, 2019.

\bibitem{fengDeepMultimodalObject2020}
Di Feng, Christian {Haase-Schuetz}, Lars Rosenbaum, Heinz Hertlein, Claudius
  Glaeser, Fabian Timm, Werner Wiesbeck, and Klaus Dietmayer.
\newblock Deep multi-modal object detection and semantic segmentation for
  autonomous driving: Datasets, methods, and challenges.
\newblock {\em arXiv:1902.07830 [cs]}, Feb. 2020.

\bibitem{girshickFastRCNN2015}
Ross Girshick.
\newblock Fast r-cnn.
\newblock {\em 2015 IEEE International Conference on Computer Vision (ICCV)},
  Dec 2015.

\bibitem{Ku_2018}
Jason Ku, Melissa Mozifian, Jungwook Lee, Ali Harakeh, and Steven~L. Waslander.
\newblock Joint 3d proposal generation and object detection from view
  aggregation.
\newblock {\em 2018 IEEE/RSJ International Conference on Intelligent Robots and
  Systems (IROS)}, Oct 2018.

\bibitem{kundu3DRCNNInstanceLevel3D2018}
Abhijit Kundu, Yin Li, and James~M. Rehg.
\newblock {{3D}}-{{RCNN}}: {{Instance}}-{{Level 3D Object Reconstruction}} via
  {{Render}}-and-{{Compare}}.
\newblock In {\em 2018 {{IEEE}}/{{CVF Conference}} on {{Computer Vision}} and
  {{Pattern Recognition}}}, pages 3559--3568, {Salt Lake City, UT, USA}, June
  2018. {IEEE}.

\bibitem{li3DFullyConvolutional2017}
Bo Li.
\newblock {{3D Fully Convolutional Network}} for {{Vehicle Detection}} in
  {{Point Cloud}}.
\newblock {\em arXiv:1611.08069 [cs]}, Jan. 2017.

\bibitem{liVehicleDetection3D2016}
Bo Li, Tianlei Zhang, and Tian Xia.
\newblock Vehicle {{Detection}} from {{3D Lidar Using Fully Convolutional
  Network}}.
\newblock {\em arXiv:1608.07916 [cs]}, Aug. 2016.

\bibitem{Liang_2019_CVPR}
Ming Liang, Bin Yang, Yun Chen, Rui Hu, and Raquel Urtasun.
\newblock Multi-task multi-sensor fusion for 3d object detection.
\newblock In {\em The IEEE Conference on Computer Vision and Pattern
  Recognition (CVPR)}, June 2019.

\bibitem{linFocalLossDense2018}
Tsung-Yi Lin, Priya Goyal, Ross Girshick, Kaiming He, and Piotr Doll{\'a}r.
\newblock Focal {{Loss}} for {{Dense Object Detection}}.
\newblock {\em arXiv:1708.02002 [cs]}, Feb. 2018.

\bibitem{Meyer_2019}
Gregory~P. Meyer, Jake Charland, Darshan Hegde, Ankit Laddha, and Carlos
  Vallespi-Gonzalez.
\newblock Sensor fusion for joint 3d object detection and semantic
  segmentation.
\newblock {\em 2019 IEEE/CVF Conference on Computer Vision and Pattern
  Recognition Workshops (CVPRW)}, Jun 2019.

\bibitem{mousavian3DBoundingBox2017a}
Arsalan Mousavian, Dragomir Anguelov, John Flynn, and Jana Kosecka.
\newblock {{3D Bounding Box Estimation Using Deep Learning}} and {{Geometry}}.
\newblock {\em arXiv:1612.00496 [cs]}, Apr. 2017.

\bibitem{mousavian20173d}
Arsalan Mousavian, Dragomir Anguelov, John Flynn, and Jana Kosecka.
\newblock 3d bounding box estimation using deep learning and geometry.
\newblock In {\em Proceedings of the IEEE Conference on Computer Vision and
  Pattern Recognition}, pages 7074--7082, 2017.

\bibitem{Nabati_2019}
Ramin Nabati and Hairong Qi.
\newblock {RRPN}: Radar region proposal network for object detection in
  autonomous vehicles.
\newblock {\em 2019 IEEE International Conference on Image Processing (ICIP)},
  Sep 2019.

\bibitem{nabatiRadarCameraSensorFusion2020}
Ramin Nabati and Hairong Qi.
\newblock Radar-{{Camera Sensor Fusion}} for {{Joint Object Detection}} and
  {{Distance Estimation}} in {{Autonomous Vehicles}}.
\newblock {\em arXiv:2009.08428 [cs]}, Sept. 2020.

\bibitem{newellHourglass2016a}
Alejandro Newell, Kaiyu Yang, and Jia Deng.
\newblock Stacked {{Hourglass Networks}} for {{Human Pose Estimation}}.
\newblock {\em arXiv:1603.06937 [cs]}, July 2016.

\bibitem{nobisDeepLearningbasedRadar2019}
Felix Nobis, Maximilian Geisslinger, Markus Weber, Johannes Betz, and Markus
  Lienkamp.
\newblock A {{Deep Learning}}-based {{Radar}} and {{Camera Sensor Fusion
  Architecture}} for {{Object Detection}}.
\newblock In {\em 2019 {{Sensor Data Fusion}}: {{Trends}}, {{Solutions}},
  {{Applications}} ({{SDF}})}, pages 1--7, Oct. 2019.

\bibitem{qiFrustumPointNets3D2018a}
Charles~R. Qi, Wei Liu, Chenxia Wu, Hao Su, and Leonidas~J. Guibas.
\newblock Frustum {{PointNets}} for {{3D Object Detection}} from {{RGB}}-{{D
  Data}}.
\newblock {\em arXiv:1711.08488 [cs]}, Apr. 2018.

\bibitem{roddick2018orthographic}
Thomas Roddick, Alex Kendall, and Roberto Cipolla.
\newblock Orthographic feature transform for monocular 3d object detection.
\newblock {\em arXiv preprint arXiv:1811.08188}, 2018.

\bibitem{shiPointRCNN3DObject2019}
Shaoshuai Shi, Xiaogang Wang, and Hongsheng Li.
\newblock {{PointRCNN}}: {{3D Object Proposal Generation}} and {{Detection}}
  from {{Point Cloud}}.
\newblock {\em arXiv:1812.04244 [cs]}, May 2019.

\bibitem{simonelli2019a}
Andrea Simonelli, Samuel Rota~Rota Bul{\`o}, Lorenzo Porzi, Manuel
  {L{\'o}pez-Antequera}, and Peter Kontschieder.
\newblock Disentangling {{Monocular 3D Object Detection}}.
\newblock {\em arXiv:1905.12365 [cs]}, May 2019.

\bibitem{wang2020infofocus}
Jun Wang, Shiyi Lan, Mingfei Gao, and Larry~S Davis.
\newblock Infofocus: 3d object detection for autonomous driving with dynamic
  information modeling.
\newblock {\em arXiv preprint arXiv:2007.08556}, 2020.

\bibitem{xuPointFusionDeepSensor2018}
Danfei Xu, Dragomir Anguelov, and Ashesh Jain.
\newblock {{PointFusion}}: {{Deep Sensor Fusion}} for {{3D Bounding Box
  Estimation}}.
\newblock {\em arXiv:1711.10871 [cs]}, Aug. 2018.

\bibitem{yanSECONDSparselyEmbedded2018}
Yan Yan, Yuxing Mao, and Bo Li.
\newblock {{SECOND}}: {{Sparsely Embedded Convolutional Detection}}.
\newblock {\em Sensors}, 18(10):3337, Oct. 2018.

\bibitem{yangRadarNetExploitingRadar2020}
Bin Yang, Runsheng Guo, Ming Liang, Sergio Casas, and Raquel Urtasun.
\newblock {{RadarNet}}: {{Exploiting Radar}} for {{Robust Perception}} of
  {{Dynamic Objects}}.
\newblock {\em arXiv:2007.14366 [cs]}, July 2020.

\bibitem{yangPIXORRealtime3D2019}
Bin Yang, Wenjie Luo, and Raquel Urtasun.
\newblock {{PIXOR}}: {{Real}}-time {{3D Object Detection}} from {{Point
  Clouds}}.
\newblock {\em arXiv:1902.06326 [cs]}, Mar. 2019.

\bibitem{yuDeepLayerAggregation2018}
Fisher Yu, Dequan Wang, Evan Shelhamer, and Trevor Darrell.
\newblock Deep {{Layer Aggregation}}.
\newblock In {\em 2018 {{IEEE}}/{{CVF Conference}} on {{Computer Vision}} and
  {{Pattern Recognition}}}, pages 2403--2412, {Salt Lake City, UT}, June 2018.
  {IEEE}.

\bibitem{7139439}
R. {Zhang}, S.~A. {Candra}, K. {Vetter}, and A. {Zakhor}.
\newblock Sensor fusion for semantic segmentation of urban scenes.
\newblock In {\em 2015 IEEE International Conference on Robotics and Automation
  (ICRA)}, pages 1850--1857, 2015.

\bibitem{zhou2019objects}
Xingyi Zhou, Dequan Wang, and Philipp Kr{\"a}henb{\"u}hl.
\newblock Objects as points.
\newblock {\em arXiv preprint arXiv:1904.07850}, 2019.

\bibitem{zhouVoxelNetEndtoEndLearning2017}
Yin Zhou and Oncel Tuzel.
\newblock {{VoxelNet}}: {{End}}-to-{{End Learning}} for {{Point Cloud Based 3D
  Object Detection}}.
\newblock {\em arXiv:1711.06396 [cs]}, Nov. 2017.

\end{thebibliography}
}

\end{document}